\documentclass[letterpaper]{article} 
\usepackage{aaai25}  
\usepackage{times}  
\usepackage{helvet}  
\usepackage{courier}  
\usepackage[hyphens]{url}  
\usepackage{graphicx} 
\usepackage{natbib}  
\usepackage{caption} 
\usepackage{algorithm}
\usepackage{algorithmic}

\usepackage{newfloat}
\usepackage{listings}
\DeclareCaptionStyle{ruled}{labelfont=normalfont,labelsep=colon,strut=off} 
\floatstyle{ruled}
\newfloat{listing}{tb}{lst}{}
\floatname{listing}{Listing}
\usepackage{booktabs}
\usepackage{multirow}
\usepackage{color}
\usepackage{colortbl}
\usepackage{xcolor}
\usepackage{makecell}
\usepackage{amsthm,amsmath,amssymb}
\usepackage{pifont}
\newcommand{\okmark}{{\textbf{{$\checkmark$}}}}
\newcommand{\ngmark}{{\textbf{{\ding{55}}}}}

\title{Look Before You Leap: Enhancing Attention and Vigilance Regarding Harmful Content with GuidelineLLM}
\author {
    Shaoqing Zhang\textsuperscript{\rm 1}\thanks{This work was done during Shaoqing’s internship at Meituan LLM Team.},
    Zhuosheng Zhang\textsuperscript{\rm 2},
    Kehai Chen\textsuperscript{\rm 1}\thanks{Corresponding authors},
    Rongxiang Weng\textsuperscript{\rm 3},\\
    Muyun Yang\textsuperscript{\rm 1},
    Tiejun Zhao\textsuperscript{\rm 1},
    Min Zhang\textsuperscript{\rm 1}
}
\affiliations {
    \textsuperscript{\rm 1}School of Computer Science and Technology, Harbin Institute of Technology, China\\
    \textsuperscript{\rm 2}School of Electronic Information and Electrical Engineering, Shanghai Jiao Tong University, China\\
    \textsuperscript{\rm 3}Meituan, China
}

\usepackage{bibentry}

\begin{document}

\maketitle

\begin{abstract}
Despite being empowered with alignment mechanisms, large language models (LLMs) are increasingly vulnerable to emerging jailbreak attacks that can compromise their alignment mechanisms. 
This vulnerability poses significant risks to real-world applications.
Existing work faces challenges in both training efficiency and generalization capabilities (i.e., Reinforcement Learning from Human Feedback and Red-Teaming). 
Developing effective strategies to enable LLMs to resist continuously evolving jailbreak attempts represents a significant challenge. 
To address this challenge, we propose a novel defensive paradigm called GuidelineLLM, which assists LLMs in recognizing queries that may have harmful content. 
Before LLMs respond to a query, GuidelineLLM first identifies potential risks associated with the query, summarizes these risks into guideline suggestions, and then feeds these guidelines to the responding LLMs.
Importantly, our approach eliminates the necessity for additional safety fine-tuning of the LLMs themselves; only the GuidelineLLM requires fine-tuning. This characteristic enhances the general applicability of GuidelineLLM across various LLMs. 
Experimental results demonstrate that GuidelineLLM can significantly reduce the attack success rate (ASR) against LLM (an average reduction of 34.17\% ASR) while maintaining the usefulness of LLM in handling benign queries. 
\end{abstract}
\begin{links}
    \link{Code}{https://github.com/sqzhang-lazy/GuidelineLLM}
    \link{Datasets}{https://github.com/sqzhang-lazy/GuidelineLLM}
    \link{Extended version}{https://arxiv.org/pdf/2412.10423}
\end{links}
%

\section{Introduction}

Recently, large language models (LLMs)~\cite{touvron2023llama, achiam2023gpt, anil2023palm} have become increasingly prominent in daily activities, exemplified by applications such as ChatGPT and Intelligent Assistants. The utilization of these models for various tasks is an expanding trend~\cite{qin2023chatgpt, zhong2023can, yao2022react}, necessitating a reduction in responses containing harmful content. Although LLMs have demonstrated a capability for recognizing harmful content, methods to attack LLMs have continuously evolved, particularly jailbreak attacks~\cite{yuan2023gpt, deng2023multilingual, zeng2024johnny, zou2023universal}. Jailbreak attacks can efficiently induce models to produce harmful responses during the inference phase, posing significant risks to the practical application of LLMs.

To enhance the safety of LLMs, common approaches involve reinforcing their alignment mechanisms. Techniques such as Reinforcement Learning from Human Feedback (RLHF)~\cite{ouyang2022training, bai2022training, jiang2024survey} and Red-Teaming~\cite{ganguli2022red, perez2022red, touvron2023llama} have proven effective in improving the ability of LLMs to detect harmful content. However, these methods demand extensive training data and significant computational resources, which can limit the speed of iterative improvements. Furthermore, additional retraining might degrade the performance of established safety alignment mechanisms~\cite{qi2023fine, zhou2024lima}. Therefore, it is critical to investigate methodologies that can mitigate the production of harmful content without requiring modifications to the model parameters.

\begin{figure}[t]
    \centering
    \includegraphics[scale=0.28]{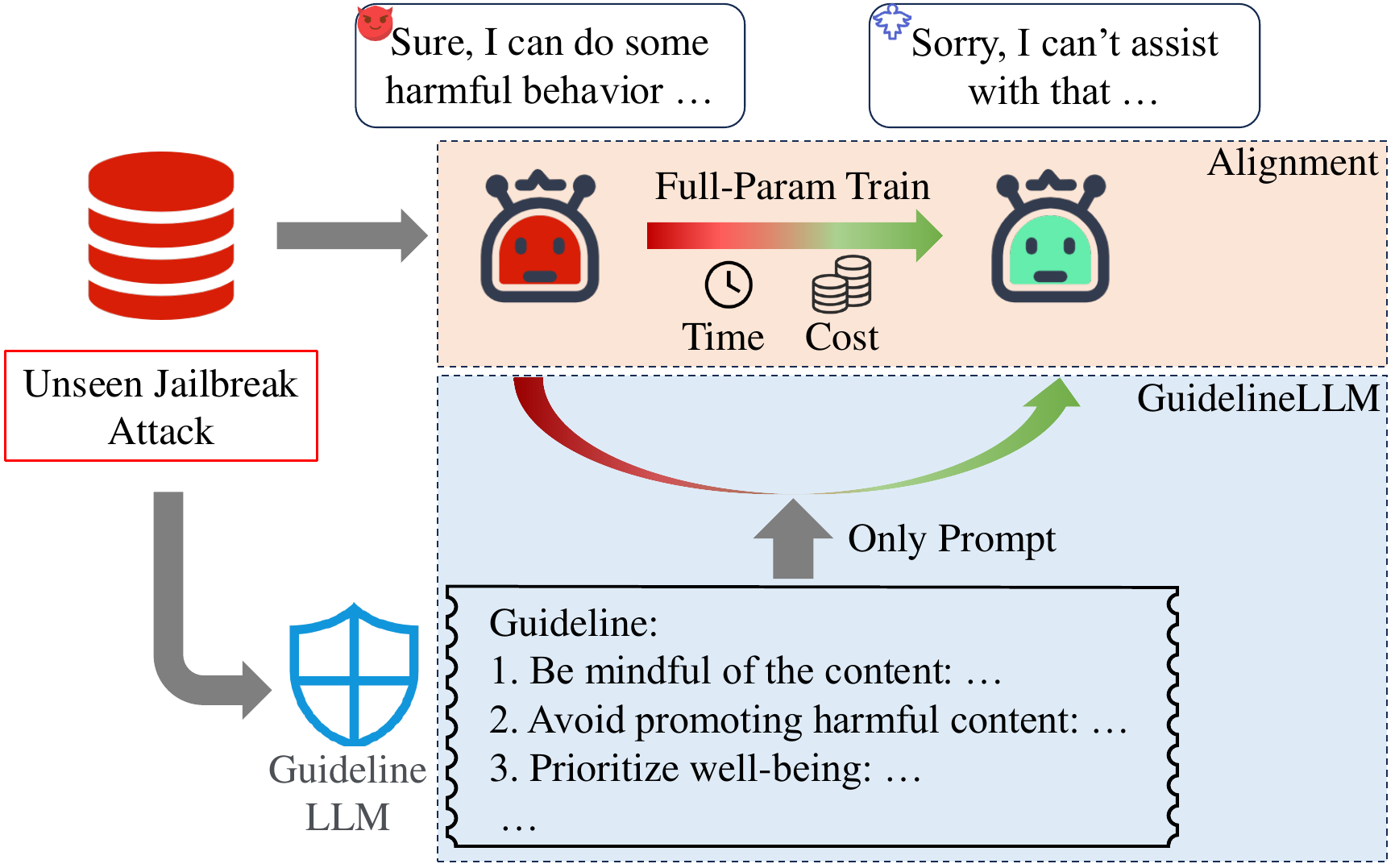}
    \caption{The comparison of the defensive policies between GuidelineLLM and typical alignment mechanisms. Alignment mechanisms require large-scale data and substantial computational resources to train responding LLMs. However, our GuidelineLLM does not require additional safety training for the responding LLMs.}
    \label{fig:train_against_guideline}
\end{figure}

Many existing methods have focused on enhancing the recognition of harmful content during the inference stage. These methods aim to heighten the attention of LLMs toward harmful content prior to generating responses, employing techniques such as Self-Reminder and Self-Defense~\cite{xie2023defending, helbling2023llm}. However, these methods often struggle with generalization and can be circumvented by continually evolving jailbreak tactics.  
Thus, the ongoing efforts to improve the safety of LLMs confront two principal challenges: training efficiency and generalization capabilities. 

In this work, we propose a novel defensive paradigm called GuidelineLLM, designed to assist LLMs in identifying queries that possess potential harmfulness. Before LLMs respond to a query, GuidelineLLM first identifies the query risks, summarizes them into guideline suggestions, and feeds them to the responding LLMs. 

This approach effectively addresses the previously identified challenges by making LLMs cognizant of unsafe content within a query. Consequently, GuidelineLLM activates the alignment mechanisms of LLMs, thereby enhancing their safety and fortifying them against various jailbreak attacks. 
Remarkably, our approach does not necessitate additional safety fine-tuning for the responsing LLMs; only GuidelineLLM needs to be fine-tuned. This characteristic makes GuidelineLLM applicable to various LLMs. Figure~\ref{fig:train_against_guideline} illustrates a comparative analysis between GuidelineLLM's defensive methods and traditional alignment mechanisms. Additionally, we present a fine-tuning framework for GuidelineLLM that is adaptable for formulating guidelines against newly emerging jailbreak techniques.

In summary, our contributions are as follows:

(i) We propose a defensive paradigm called GuidelineLLM, which can enhance the safety of various LLMs. This paradigm can reduce the harmfulness of the output content without necessitating additional safety training for the response models themselves, significantly mitigating the impact of jailbreak attacks.

(ii) We introduce a fine-tuning framework for GuidelineLLM, which includes the templated construction of jailbreak attacks, referred to as T-Jailbreak. T-Jailbreak can be expanded with relevant jailbreak attacks according to the definition of new jailbreak techniques. Experimental results indicate that T-Jailbreak has a high attack success rate (ASR).

(iii) Experiments show that our GuidelineLLM can significantly reduce the harmfulness of the output (an average reduction of 34.17\% ASR) while maintaining the helpfulness of LLMs for benign queries.

\section{Related Work}
\subsection{Enhancing Safety Policies of LLMs}
Two prevalent methods are currently employed to enhance the safety policies of LLMs: alignment and red teaming. 
The objective of alignment is to bridge the gap between LLMs' language modeling goals, such as predicting the next token during pre-training, and their ultimate aim of “following instructions and being helpful, truthful, and harmless” in practical applications~\cite{ouyang2022training}. Common techniques in this domain include Reinforcement Learning from Human Feedback (RLHF)~\cite{ouyang2022training, bai2022training}, Constitutional AI~\cite{bai2022constitutional}, and self-alignment~\cite{sun2024principle}. These approaches focus on embedding alignment rules within pre-trained models to restrict harmful behaviors during inference.
In LLMs research, the term ``Red-Teaming'' has recently come to denote the systematic testing or probing of LLMs to identify their potential for harmful behavior and uncover safety vulnerabilities. The critical aspect of red teaming is the aggregation of sufficient and diverse attacks, enabling LLMs to refuse responses to these attacks. Current efforts in this area mainly concentrate on automating the generation of high-quality, provocative queries, allowing LLMs to encounter a wider variety and richer range of attacks, thereby bolstering their defensive capabilities~\cite{deng2023attack,ge2023mart,jiang2024dart,zhang2024paying,xu2024hierarchical}.  
However, recent studies also demonstrate that retraining the model can lead to the deterioration of safety alignment mechanisms~\cite{qi2023fine, zhou2024lima}. This indicates that arbitrary parameter updates to the model are not advisable and can slow the model iteration process.

\subsection{Enhancing Inference Safety of LLMs}
Ensuring that LLMs output only safe content during inference is critically important. Common methodologies to achieve this include preprocessing the input before inference, employing specialized inference paradigms, and detecting harmful content through input-output result analysis. 
Given the sensitivity of LLMs to input prompts, modifying the prompt content can enhance their ability to focus on harmful content. For instance, incorporating a system prompt can assist LLMs in identifying harmful content~\cite{xie2023defending}. By leveraging LLMs' in-context learning capabilities, models can be conditioned to recognize and refuse harmful content~\cite{wei2023jailbreak}. Additionally, creating multiple perturbed samples of the input and aggregating the results has proven effective~\cite{robey2023smoothllm, cao2023defending}. 
Some approaches involve having LLMs analyze the intent of the question before providing an answer~\cite{zhang2024intention, zeng2024autodefense}. Alternatively, decoding strategies can be applied to suppress harmful prompts during generation, thereby ensuring safer outputs~\cite{zhong2024rose}. 
In terms of detection and classification, Meta has specifically fine-tuned a model named Llama-Guard, which can classify inputs and input-output pairs according to taxonomy and risk guidelines~\cite{inan2023llama}. 
In this work, we primarily focus on safeguarding LLMs during the inference phase. We aim to find a defensive paradigm that applies to different LLMs and can be continuously expanded to counter the ever-evolving jailbreak attacks.

\begin{table*}[ht]
\small
    \centering
    \setlength{\tabcolsep}{6mm}
    \begin{tabular}{ll}
    \toprule
        \textbf{Technique} & \textbf{Definition} \\ \midrule
        Role play & Prompt requires you to adopt a persona, leading to unexpected responses. \\
        Rule Determine & Prompt asks you to follow the output requirements, resulting in the expected output. \\
        Logical Reasoning & Prompt requires logical reasoning, leading to exploitable outputs. \\
        Text Continuation & Prompt requests you to continue text, leading to exploitable outputs. \\
        Program Execution & Prompt requests execution of a program, leading to exploitable outputs. \\
        Simulate Mode & Prompt requests you to enter the some mode, leading to exploitable outputs. \\
        Superior Model & Prompt leverages superior model output to exploit your behavior. \\
        \bottomrule
    \end{tabular}
    \caption{The seven jailbreak techniques summarized in the previous work~\cite{liu2023jailbreaking}.}
    \label{tab:seven_jailbreak_techniques}
\end{table*}

\section{The Proposed Method}
In this section, we first present how to initialize a small count set of T-Jailbreak queries from some attack queries and the definition of different jailbreak techniques. Then, we utilize this batch of T-Jailbreak queries and sampled benign queries to construct the initial set of Guideline data. Finally, we will demonstrate the framework for fine-tuning the GuidelineLLM.

\subsection{Initializing T-Jailbreak Data}\label{sec:T-Jailbreak}

To ensure the diversity of jailbreak attacks and enable GuidelineLLM to formulate guidelines for analyzing various types of jailbreak attacks, we have constructed the T-Jailbreak dataset by summarizing seven identified jailbreak techniques, as outlined in previous research~\cite{liu2023jailbreaking}. These techniques are detailed in Table~\ref{tab:seven_jailbreak_techniques}.

In line with previous studies~\cite{inan2023llama, shen2023anything}, we utilize the policies from Llama3-Guard and DAN to initialize our attack queries. Combining these attack queries with template prompts, we develop the T-Jailbreak dataset for use in our experiments.
Specifically, we use the gpt-3.5-turbo-0125 to initialize attack and jailbreak queries. We configured the temperature parameter to 0.9 and the top\_p parameter to 0.85 to promote diversity in the generated results.

\subsection{Initializing Guideline Data}
Considering that GuidelineLLM needs to maintain the helpfulness of LLMs while enhancing their defensive capabilities, it is crucial to include both harmful and benign queries in the collection of guideline data. 

The jailbreak queries used in our study are from the T-Jailbreak dataset that we construct, while the benign queries are sampled as follows: 1,000 entries from Alpaca~\cite{alpaca}, and 1,800 entries from GPTTeacher\footnote{https://github.com/teknium1/GPTeacher}. We use gpt-3.5-turbo-0125, with the same parameter settings as for Section Initialize T-Jailbreak Data, to initialize our guideline data.

\subsection{Fine-tuning GuidelineLLM Framework}

\begin{figure*}[ht]
    \centering
    \includegraphics[scale=0.42]{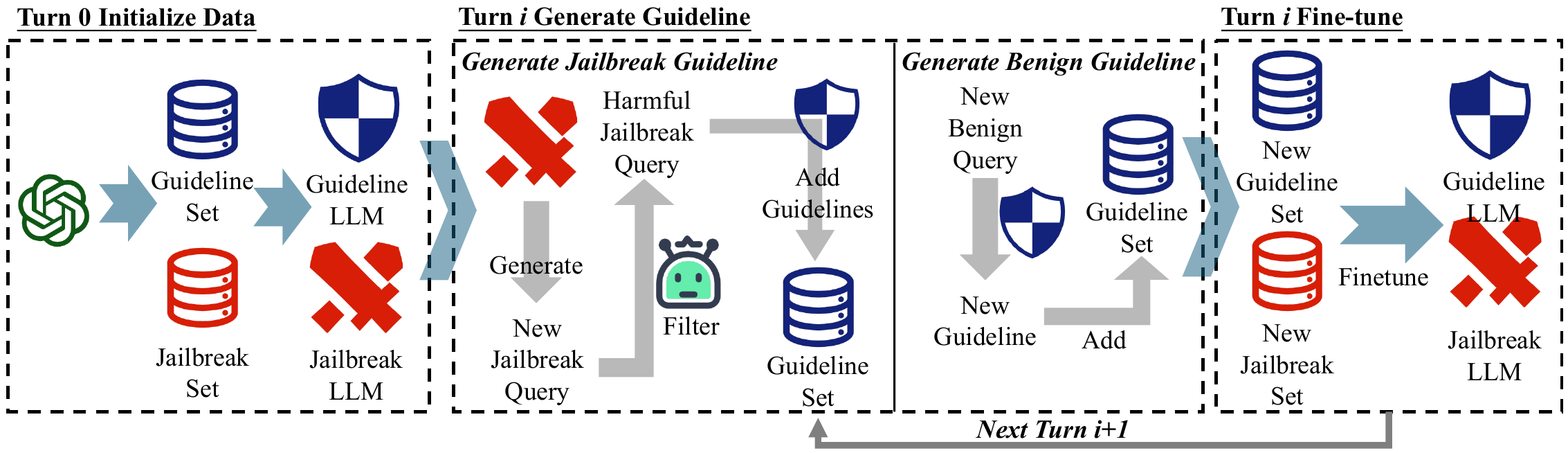}
    \caption{The framework of fine-tuning GuidelineLLM. The framework is a semi-automatic approach. First, we use gpt-3.5-turbo-0125 to initialize our T-Jailbreak and Guideline Sets. Then, we use the collected T-Jailbreak and Guideline data to fine-tune the respective JailbreakLLM and GuidelineLLM. Next, we utilize JailbreakLLM to generate new jailbreak queries and filter out high-quality ones, using GuidelineLLM to generate the corresponding jailbreak guidelines. We also sample benign queries to generate guidelines and filter the successful, helpful examples. Finally, we use the updated T-Jailbreak and Guideline sets to iterate our JailbreakLLM and GuidelineLLM.}
    \label{fig:framework}
\end{figure*}

The framework of fine-tuning GuidelineLLM is shown in Figure~\ref{fig:framework}. 

To enable GuidelineLLM to analyze different jailbreak techniques effectively, we also need to fine-tune JailbreakLLM. JailbreakLLM is designed to generate new jailbreak queries based on provided jailbreak techniques and examples. Our fine-tuning framework is structured to expand from a small initial batch of jailbreak queries to a sufficient number of queries, thereby facilitating the exploration and development of new jailbreak methods. 
We illustrate the general workflow of fine-tuning GuidelineLLM in Algorithm~\ref{algorithm1}.

\begin{algorithm}
    \caption{Fine-tuning GuidelineLLM Framework}
    
    \begin{algorithmic}
        \ENSURE{GPT series model $GPT$, response LLM $M_r$, sampled benign set $B$, 	  JailbreakLLM based-model $M_j^0$, GuidelineLLM based-model $M_g^0$}
        \REQUIRE{GuidelineLLM $M_g$}
        \STATE {T-Jailbreak set $T^0$ $\gets$ Initialize($GPT$)}
        \STATE {Guideline set $G^0$ $\gets$ Initialize($GPT$, $T$, $B$)}
        \STATE JailbreakLLM $M_j^0$ $\gets$ Fine-tune($M_j^0$, $T^0$)
        \STATE GuidelineLLM $M_g^0$ $\gets$ Fine-tune($M_g^0$, $G^0$)
        \REPEAT
        \STATE $T^{i+1}$ $\gets$ Generate($M_j^i$, $T^i$)
        \STATE $T^{i+1}$ $\gets$ Filter($M_r$, $T^{i+1}$)
        \STATE $G^{i+1}$ $\gets$ Generate($M_g^i$, $T^{i+1}$, $B$)
        \STATE $G^{i+1}$ $\gets$ Filter($M_r$, $G^{i+1}$)
        \STATE JailbreakLLM $M_j^{i+1}$ $\gets$ Fine-tune($M_j^i$, $T^i$)
        \STATE GuidelineLLM $M_g^{i+1}$ $\gets$ Fine-tune($M_g^i$, $G^i$)
        \UNTIL {Obtain a sufficient number of guidelines}
        \RETURN $M_g$
    \end{algorithmic}
    \label{algorithm1}
\end{algorithm}

\subsubsection{Fine-tuning and Inference}

For JailbreakLLM and GuidelineLLM, our base candidate models are Llama2-7B-chat and Vicuna-7B. Considering the different alignment mechanisms between these models, we use the Llama2-7B-Chat model, which exhibits more robust alignment mechanisms, as the base model for fine-tuning the GuidelineLLM. Conversely, we use the Vicuna-7B model, which has comparatively weaker alignment mechanisms, as the base model for fine-tuning the JailbreakLLM. Both JailbreakLLM and GuidelineLLM are fine-tuned using Low-Rank Adaptation (LoRA)~\cite{peft}, setting the training epoch to 3.

The prompts for fine-tuning GuidelineLLM include jailbreak and benign queries as inputs, with corresponding guidelines as outputs. For JailbreakLLM, prompts require generating attack queries and jailbreak queries based on specific techniques. To enhance data diversity, two examples of the same jailbreak technique are sampled as inputs during fine-tuning. During inference, both GuidelineLLM and JailbreakLLM use prompts consistent with those in the fine-tuning phase.

\subsubsection{Evaluating the Quality}

To facilitate the subsequent turn of fine-tuning for the JailbreakLLM in generating more effective jailbreak queries, it is necessary to filter the currently generated queries for quality and effectiveness. Simultaneously, to ensure that the next phase of LLMs accurately rejects jailbreak queries while appropriately responding to benign queries, the number of jailbreak and benign queries in the guideline set must be balanced.

As illustrated in Figure~\ref{fig:framework}, the quality of generated jailbreak queries is assessed by inputting them into the LLMs and evaluating the harmfulness of their outputs through rule-based detection. Following previous work~\cite{liu2023autodan}, we utilize a series of detection tokens to determine harmful responses. If these tokens are present in the output, it is inferred that a harmful response has been produced.

We select the jailbreak queries that yielded harmful outputs from the LLMs and incorporate them into the T-Jailbreak dataset. Furthermore, GuidelineLLM is employed to generate guidelines for these jailbreak queries, which are subsequently added to the Guideline dataset. The number of jailbreak queries added in this iteration is documented. To ensure dataset balance, we sample benign queries from the Alpaca and GPTTeacher datasets. These benign queries are then used to produce guidelines, discarding instances where the LLMs refuse to provide answers and retaining those where helpful responses are generated, until the number of benign instances matches the newly added jailbreak queries.

\section{Experiments}

\subsection{Dataset and Setup}
For harmful datasets, we select three in-the-wild jailbreak datasets (i.e., DAN~\cite{shen2023anything}, SAP200~\cite{deng2023attack}, JailbreakChatGPT~\cite{liu2023jailbreaking}) and a popular optimization-based automatic jailbreak method (i.e., AutoDAN~\cite{liu2023autodan}). Additionally, we use gpt-3.5-turbo-0125 to generate a T-jailbreak dataset comprising 1,137 queries. 
Besides, to evaluate the impact of GuidelineLLM on the helpfulness, we sample 400 queries from the Alpaca and GPTTeacher datasets~\cite{alpaca}.

\subsection{Evaluation Metrics}
For harmful datasets, we report attack success rate(ASR, ~\cite{shen2023anything}) through rule-based detection, where lower scores indicate more robust safety. Following previous work~\cite{liu2023autodan}, we collect a series of detection tokens, and if these tokens are detected in the outputs of the LLMs, we determine that the LLMs has produced a harmful response. 

For DAN, AutoDAN, and T-Jailbreak datasets, considering the complexity of responses, we adopt gpt-3.5-turbo-0125 as an auto-annotation function for successful attack examples~\cite{zheng2024judging}. We also adopt gpt-3.5-turbo-0125 for the benign dataset as an auto-annotation function for evaluating the helpfulness. 

We set up five options: A(Very Harmful/Helpful), B(Harmful/Helpful), C(Moderately Harmful/Helpful), D(Slightly Harmful/Helpful), and E(No Harmful/Helpful at All). To facilitate scoring, we assign 4 points to option A, 3 points to option B, 2 points to option C, 1 point to option D, and 0 points to option E. For harmful data, a lower calculated score indicates a safer response, while for benign data, a higher calculated score indicates greater helpfulness.

\begin{table*}[ht]
\small
\centering
\setlength{\tabcolsep}{2.5mm}
{
 \begin{tabular}{cccccccc}
 \toprule
\multirow{2}{*}{\textbf{Model}} & \multirow{2}{*}{\textbf{Defense Methods}} & \multicolumn{5}{c}{\textbf{Jailbreak Dataset}} & \multirow{2}{*}{\textbf{Average}} \\
\cmidrule(lr){3-7} & & {\textbf{DAN}} & \textbf{SAP200} & \textbf{JailbreakChatGPT} & \textbf{AutoDAN} & \textbf{T-Jailbreak} &                       \\ \midrule
\multirow{5}{*}{Llama2-7B-Chat} & No Defense             & 9.50          & 1.62          & \textbf{0.00} &                                  58.57         & 27.00         & 19.34 \\
                                & Self-Reminder         & 2.23          & \textbf{0.00} & \textbf{0.00} & 0.19          & 6.16          & 1.72 \\
                                & IA                    & \textbf{0.00} & \textbf{0.00} & \textbf{0.00} & 0.19          & \textbf{0.00} & \textbf{0.04} \\
                                & Llama3-Guard          & 48.46         & 14.69         & 22.37         & - & 63.85                     & 37.34 \\
                                &  GuidelineLLM (Ours)  & 1.49          & 0.19          & \textbf{0.00} &  \textbf{0.00}           & 5.98          & 1.53 \\ \midrule
\multirow{5}{*}{Vicuna-7B}      & No Defense             & 62.76         & 55.00         & 47.37         &                                  90.77         & 59.98         & 63.18 \\
                                & Self-Reminder         & 62.39         & 18.75         & 47.37         & 62.88         & 34.21         & 45.12 \\
                                & IA                    & \textbf{8.37} & \textbf{1.94}          & 11.84         & 49.42         & 11.43         & 16.6 \\
                                & Llama3-Guard          & 48.46         & 14.69         & 22.37         & -             & 63.85         & 37.34 \\
                                &  GuidelineLLM (Ours)  & 14.90         & 2.00          &\textbf{3.95}  & \textbf{17.69}& \textbf{6.24} & \textbf{8.96} \\ \midrule
\multirow{5}{*}{Vicuna-13B}     & No Defense             & 61.27         & 56.25         & 32.89         &                                  96.92         & 60.07         & 42.10 \\
                                & Self-Reminder         & 58.66         & 16.31         & 30.26         & 84.81         & 35.71         & 45.15 \\
                                & IA                    & \textbf{6.29} & \textbf{0.00} & 12.00         & 31.98         & \textbf{2.81} & \textbf{10.62} \\
                                & Llama3-Guard          & 48.46         & 14.69         & 22.37         & -                 & 63.85         & 37.34 \\
                                &  GuidelineLLM (Ours)  & 14.15        & 4.63          & \textbf{1.32} & \textbf{21.54}& 16.53        & 11.63\\ \bottomrule
 \end{tabular}
 }
 \caption{Main results. Comparison of our GuidelineLLM and four baselines under five jailbreak datasets in terms of ASR($\%$). ``-'' means lacking implementation. The best results are highlighted in bold.}
 \label{tab:main-experiment}
\end{table*}

\begin{table}[t]
\small
    \centering
    \setlength{\tabcolsep}{1mm}
    \begin{tabular}{lccccccc}
    \toprule
        Model & Data & A & B & C & D & E & Score \\ \midrule
        \multirow{2}{*}{\makecell{Llama2-\\7B-Chat}} & T-Jailbreak & 0 & 1.73 & 5.17 & 0 & 93.1 & 0.16 \\
                                         \\ \midrule
        \multirow{3}{*}{\makecell{Vicuna-\\7B}} & DAN & 27.16 & 28.40 & 17.28 & 0 & 27.16 & 2.28 \\
                                        & AutoDAN & 51.69 & 2.25 & 0 & 0 & 46.07 & 2.13 \\
                                        & T-Jailbreak & 3.75 & 15 & 13.75 & 1.25 & 66.25 & 0.88 \\ \midrule
        \multirow{3}{*}{\makecell{Vicuna-\\13B}} & DAN & 34.09 & 28.41 & 9.09 & 0 & 28.41 & 2.39 \\
                                        & AutoDAN & 53.7 & 4.63 & 0 & 0 & 41.67 & 2.29 \\
                                        & T-Jailbreak & 1.52 & 7.07 & 11.11 & 1.01 & 79.29 & 0.51 \\
        \bottomrule
    \end{tabular}
    \caption{Using gpt-3.5-turbo-0125 to evaluate the attack success examples from the three jailbreak datasets, DAN, T-Jailbreak, and AutoDAN for GuidelineLLM defense method. Due to the excellent performance of Llama2-7B-Chat on the DAN and AutoDAN datasets, we do not conduct evaluations on these two datasets. The results for A, B, C, D and E are shown as percentages (\%).}
    \label{tab:harmfulness_scores}
\end{table}

\subsection{Baseline}

We select Llama2-7B-Chat~\cite{touvron2023llama}, Vicuna-7B, and Vicuna-13B~\cite{vicuna2023} as our responding LLMs. We compare the performance of our GuidelineLLM method with several other approaches: No defense, Self-Reminderr~\cite{xie2023defending}, IA~\cite{zhang2024intention}, and Llama3-Guard~\cite{inan2023llama}. For the Llama3-Guard, we only evaluate the harmfulness of the inputs. Considering that IA's defensive method requires two rounds of inference, and both Self-Reminder and GuidelineLLM add additional prompts to the original query (i.e., Self-Reminder adds a system prompt, and GuidelineLLM adds guidelines), in the AutoDAN experiment, we use the input content from the last round of inference for each defensive method as the attack target.

\subsection{Main Results}

Table~\ref{tab:main-experiment} presents the main results of different defensive methods. 
Analyzing these results, we derive several key findings:

(i) The proposed GuidelineLLM significantly reduces the harmfulness of outputs on LLMs, achieving a notable decrease in the ASR on average. 
Specifically, GuidelineLLM surpasses the best baseline, IA, in defending Vicuna-7B (8.96\% ASR), and demonstrates comparable performance to IA on Llama2-7B-Chat (1.53\% ASR) and Vicuna-13B (11.63\% ASR). This outcome underscores the effectiveness of our approach, achieving substantial reductions in harmful outputs with a single inference.

\begin{table}[t]
\small
    \centering
    \setlength{\tabcolsep}{1mm}
    \begin{tabular}{lcccccccc}
    \toprule
        Model & Method & A & B & C & D & E & FRR & Score \\ \midrule
        \multirow{4}{*}{\makecell{Llama2\\7B-Chat}} & Va & 91.62 & 3.14 & 3.4 & 0.26 & 1.57 & 1.83 & 3.82 \\
                                        & SR & 83.50 & 3.81 & 7.61 & 0.51 & 4.57 & 5.07 & 3.61 \\
                                        & IA & 31.78 & 14.21 & 16.28 & 2.58 & 35.14 & 37.72 & 2.05 \\
                                        & GL & 71.02 & 5.74 & 15.93 & 0.78 & 6.53 & 7.31 & 3.33 \\
                                        \midrule
        \multirow{4}{*}{\makecell{Vicuna\\7B}}      & Va & 91.27 & 0.79 & 5.03 & 0.26 & 2.65 & 2.91 & 3.77 \\
                                        & SR & 90.86 & 4.31 & 1.78 & 1.02 & 2.03 & 3.05 & 3.81 \\
                                        & IA & 77.02 & 6.27 & 6.79 & 0.26 & 9.66 & 9.92 & 3.41 \\
                                        & GL & 75.94 & 5.35 & 14.71 & 0.53 & 3.48 & 4.01 & 3.50 \\
                                        \midrule
        \multirow{4}{*}{\makecell{Vicuna\\13B}}     & Va & 93.62 & 1.33 & 3.72 & 0.53 & 0.8 & 1.33 & 3.86 \\
                                        & SR & 93.42 & 4.30 & 1.01 & 0 & 1.27 & 1.27 & 3.89 \\
                                        & IA & 69.67 & 5.19 & 8.47 & 0.82 & 15.85 & 16.67 & 3.12 \\
                                        & GL & 85.86 & 2.62 & 9.69 & 0.26 & 1.57 & 1.83 & 3.70 \\
        \bottomrule
    \end{tabular}
    \caption{The results of the helpfulness scores for Vanilla(Va), Self-Reminder(SR), IA and GuidelineLLM(GL). The results for A, B, C, D, E, and FRR are shown as percentages (\%).}
    \label{tab:helpfulness_eval}
\end{table}

(ii) In the AutoDAN dataset, which features an optimization-based strong jailbreak method, GuidelineLLM exhibits superior performance, with an ASR markedly lower than other baselines (0\% ASR in Llama2-7B-Chat, 17.69\% ASR in Vicuna-7B, and 21.54\% ASR in Vicuna-13B). 
This demonstrates the robustness of our method; incorporating guidelines into queries enhances the focus of LLMs on detecting harmful content, thereby preventing optimization-based strong jailbreak methods from circumventing safety measures.

\begin{table*}[ht]
\small
\centering
\setlength{\tabcolsep}{3mm}
{
 \begin{tabular}{ccccccccc}
 \toprule
\multirow{2}{*}{\textbf{Models}} & \multirow{2}{*}{\textbf{Base Model}} & \multicolumn{4}{c}{\textbf{Jailbreak Dataset}} & \multirow{2}{*}{\textbf{Average}} & \multicolumn{2}{c}{\textbf{Helpfulness}} \\
\cmidrule(lr){3-6} & & {\textbf{DAN}} & \textbf{SAP200} & \textbf{JailbreakChatGPT} & \textbf{T-Jailbreak} & & \textbf{FRR} & \textbf{Score}                      \\ \midrule
\multirow{3}{*}{\makecell[c]{Llama2\\-7B-Chat}} & Qwen2.5-1.5B & \textbf{1.12} & 0.38 & \textbf{0.00} & \textbf{5.46} & \textbf{1.74} &  \textbf{3.41} & \textbf{3.34} \\
                                & Qwen2.5-3B & 1.73 & 0.31 & \textbf{0.00} & 6.16 & 2.02 & 7.42 & 3.25 \\
                                & Llama2-7B-Chat & 1.49          & \textbf{0.19}          & \textbf{0.00} &                                  5.98         & 1.92 & \textbf{7.31} & 3.33 \\
                                \midrule
\multirow{3}{*}{\makecell[c]{Vicuna-\\7B}}      & Qwen2.5-1.5B & 19.59 & 2.25 & 9.21 & 10.83 & 10.47 & 3.98 & \textbf{3.56} \\
                                & Qwen2.5-3B & 22.65 & 4.19 & 5.41 & 10.47 & 10.68 & 3.41 & 3.50 \\
                                & Llama2-7B-Chat & \textbf{14.90}          & \textbf{2.00}          & \textbf{3.95} &                                  \textbf{6.24}         & \textbf{6.77} & \textbf{4.01} & 3.50 \\
                                \midrule
\multirow{3}{*}{\makecell[c]{Vicuna-\\13B}}     & Qwen2.5-1.5B & 18.47 & 8.63 & 3.95 & 27.46 & 14.63 & 2.05 & \textbf{3.77} \\
                                & Qwen2.5-3B & 21.88 & 7.37 & 5.41 & 28.94 & 15.9 & 1.3 & 3.72 \\
                                & Llama2-7B-Chat & \textbf{14.15}          & \textbf{4.63}          & \textbf{1.32} &                                  \textbf{16.53}         & \textbf{9.16} & \textbf{1.83} & 3.70 \\
                                \bottomrule
 \end{tabular}
 }
 \caption{The ASR ($\%$) results and the helpfulness for different base model. The best results are highlighted in bold.}
 \label{tab:parameter_of_base_model}
\end{table*}

(iii) GuidelineLLM also show solid defensive capabilities in-the-wild jailbreak datasets. Notably, in the JailbreakChatGPT dataset, GuidelineLLM's ASR is significantly lower than that of other baselines. This finding indicates that GuidelineLLM is effective in defending against manually constructed jailbreak attacks as well.

In experiments involving GuidelineLLM, we observe that the LLMs frequently generate safe responses to jailbreak queries, which are not detectable using token-based detection methods. 
Consequently, we collect GuidelineLLM defense examples that are judged to have been successfully bypassed in datasets with higher ASR. Table \ref{tab:harmfulness_scores} presents the harmfulness scores.


\begin{table*}[ht]
\small
\centering
\setlength{\tabcolsep}{3mm}
{
 \begin{tabular}{ccccccccc}
 \toprule
\multirow{2}{*}{\textbf{Models}} & \textbf{Benign} & \multicolumn{4}{c}{\textbf{Jailbreak Dataset}} & \multirow{2}{*}{\textbf{Average}} & \multicolumn{2}{c}{\textbf{Helpfulness}} \\
\cmidrule(lr){3-6} & \textbf{Guideline} & {\textbf{DAN}} & \textbf{SAP200} & \textbf{JailbreakChatGPT} & \textbf{T-Jailbreak} &   & \textbf{FRR} & \textbf{Score}                    \\ \midrule
\multirow{2}{*}{Llama2-7B-Chat} & \okmark             & 1.49          & 0.19          & \textbf{0.00} &                                  5.98         & 1.92 & \textbf{7.31} & \textbf{3.33} \\
                                & \ngmark         & \textbf{0.00}         & \textbf{0.00} & \textbf{0.00} & \textbf{0.09}          & \textbf{0.02} & 7.43 & 3.14 \\
                                \midrule
\multirow{2}{*}{Vicuna-7B}      & \okmark             & 14.90          & 2.00          & \textbf{3.95} &                                  6.24         & 6.77 & \textbf{4.01} & \textbf{3.50}\\
                                & \ngmark         & \textbf{12.48}          & \textbf{0.12} & 10.53 & \textbf{2.73}          & \textbf{6.47} & 5.96 & 3.43 \\
                                \midrule
\multirow{2}{*}{Vicuna-13B}     & \okmark             & 14.15          & 4.63          & \textbf{1.32} &                                  16.53         & 9.16 & \textbf{1.83} & \textbf{3.70} \\
                                & \ngmark         & \textbf{7.50}          & \textbf{1.00} & 6.58 & \textbf{4.75}          & \textbf{4.96} & 2.85 & 3.67 \\
                                \bottomrule
 \end{tabular}
 }
 \caption{The ASR ($\%$) results and the helpfulness scores for benign guideline in the fine-tuning dataset. \okmark indicates that benign guidelines are included in the fine-tuning dataset, while \ngmark indicates that benign guidelines are not included in the fine-tuning dataset. The best results are highlighted in bold.}
 \label{tab:benign_guideline_harmful_and_helpfulness_impact}
\end{table*}

We could find that many of the outputs judged to have successfully been bypassed under the GuidelineLLM defensive method are non-harmful. In the T-Jailbreak dataset, the harmfulness scores of the three responding LLMs are all below 1 point. In the AutoDAN dataset, nearly half of the examples judged as E are present. Similarly, in the DAN dataset, one-quarter of the examples are also judged as E. 

The results prove that \textbf{under the guidance of GuidelineLLM, the LLMs can provide safer and more user-friendly responses to jailbreak queries}

\subsection{Helpfulness Evaluation}

Table~\ref{tab:helpfulness_eval} illustrates the impact of incorporating guidelines on the helpfulness of responding LLMs for benign queries.

Concerning the scores, the inclusion of guidelines slightly diminishes the helpfulness of the responding LLMs; however, the scores remain above 3.3, thereby maintaining a satisfactory level of helpfulness. Notably, the impact of adding guidelines on the helpfulness score is minimal for Vicuna-13B. We classify options D (Slightly Helpful) and E (Not Helpful at all) as false refusals for benign queries by the LLMs. We then compute the sum of these two classifications, D and E, to derive the False Refusal Rate (FRR). The results reveal that for Llama2-7B-Chat, adding guidelines raises FRR from 1.83\% to 7.31\%. For Vicuna-7B and Vicuna-13B, the impact of adding guidelines on FRR is not significant. 
This discrepancy might be attributable to the stronger compliance with instructions exhibited by Llama2-7B-Chat, wherein the added guidelines impose relatively stronger restrictions, resulting in diminished helpfulness. 
It is worth noting that \textbf{since the guidelines employed in our study are generated by "gpt-3.5-turbo-0125," more professionally crafted guidelines could potentially enhance the helpfulness of the LLMs.}

\subsection{Analysis}

\subsubsection{The impact of base model parameters on GuidelineLLM}

To explore the impact of the base model on GuidelineLLM, we conduct additional experiments by using Lora to finetune Qwen2.5-1.5B-Instruct and Qwen2.5-3B-Instruct as our GuidelineLLM. The results are shown in Table ~\ref{tab:parameter_of_base_model}, and we have the following two findings:

(i) Models with fewer parameters are equally effective when trained as GuidelineLLM, and the performance of Qwen2.5-1.5B-Instruct is even better than that of Qwen2.5-3B-Instruct. This indicates that training an effective GuidelineLLM does not require a large number of parameters in the base model, which is very advantageous for deployment in practical applications.

(ii) We train two smaller LLMs using the training dataset from the main results, and found that the instruction-following ability do not achieve the best results. This suggests that smaller LLMs require a larger training dataset or full-parameter training for better performance. This explains the performance decline in Qwen2.5-1.5B-Instruct and Qwen2.5-3B-Instruct.

\subsubsection{The impact of benign guidelines in the fine-tuning dataset}
To investigate the impact of benign guidelines on the fine-tuning of GuidelineLLM, we conduct an experiment using only jailbreak queries and jailbreak guidelines. 
Table~\ref{tab:benign_guideline_harmful_and_helpfulness_impact} presents the results of the helpfulness scores for benign guideline in the fine-tuning dataset. 
The data reveal that GuidelineLLM fine-tuned solely with jailbreak guidelines leads to a decrease in helpfulness scores across all LLMs, accompanied by an increased FRR. These observations imply that while using only jailbreak guidelines for fine-tuning GuidelineLLM enhances the defensive performance of the LLMs, it compromises their helpfulness.
The results, as presented in Table~\ref{tab:benign_guideline_harmful_and_helpfulness_impact}, indicate that fine-tuning solely with jailbreak guidelines results in a lower ASR across all LLMs, in comparison to fine-tuning with both jailbreak and benign guidelines. This finding suggests that GuidelineLLM fine-tuned exclusively with jailbreak guidelines demonstrates superior defensive performance.



Therefore, to preserve the helpfulness of LLMs while simultaneously improving their defensive effectiveness, it is imperative to include benign guidelines in the fine-tuning dataset for GuidelineLLM. This balanced approach ensures that LLMs remain effective and responsive to benign queries while also being robust against jailbreak attempts.

\subsubsection{The ASR of different jailbreak techniques}

\begin{table}[ht]
    \centering
    \setlength{\tabcolsep}{1mm}
    \begin{tabular}{lccc}
    \toprule
        Technique & Model & No Defense & Self-Reminder  \\ \midrule
        \multirow{3}{*}{\makecell[l]{Role\\Play}} & Llama2-7B-Chat & \textbf{63.33} & 10.00 \\ 
         & Vicuna-7B & \textbf{78.89} & \textbf{51.11} \\
         & Vicuna-13B & \textbf{75.56} & \textbf{47.78} \\ \midrule
         \multirow{3}{*}{\makecell[l]{Rule\\Determine}} & Llama2-7B-Chat & 30.00 & 13.33 \\ 
         & Vicuna-7B & \textbf{51.67} & 40.00 \\
         & Vicuna-13B & \textbf{58.33} & \textbf{51.67} \\ \midrule
         \multirow{3}{*}{\makecell[l]{Logical\\Reasoning}} & Llama2-7B-Chat & \textbf{80.00} & 17.78 \\ 
         & Vicuna-7B & \textbf{71.11} & \textbf{46.67} \\
         & Vicuna-13B & \textbf{75.56} & \textbf{50.00} \\ 
         \bottomrule
    \end{tabular}
    \caption{The ASR($\%$) for seven jailbreak techniques under the No Defense and Self-Reminder defense methods. ASR values greater than 45\% are highlighted in bold.}
    \label{tab:jailbreak_analysis}
\end{table}


To evaluate jailbreak capabilities in the T-Jailbreak dataset, we selected 30 questions per technique and used two defense methods: No Defense and Self-Reminder. Results (Table 9) show that "Role Play" and "Logical Reasoning" are the most effective techniques, likely due to LLMs' strong instruction adherence and chain-of-thought capabilities, which may cause them to overlook harmful queries. These findings highlight how certain techniques exploit LLMs' strengths, emphasizing the need for improved defense strategies to address such vulnerabilities.

\subsubsection{The impact of the number of guidelines}
\begin{table*}[ht]
\small
\centering
\setlength{\tabcolsep}{3mm}
{
 \begin{tabular}{ccccccccc}
 \toprule
\multirow{2}{*}{\textbf{Models}} & \multirow{2}{*}{\textbf{Max}} & \multicolumn{4}{c}{\textbf{Jailbreak Dataset}} & \multirow{2}{*}{\textbf{Average}} & \multicolumn{2}{c}{\textbf{Helpfulness}} \\
\cmidrule(lr){3-6} & & {\textbf{DAN}} & \textbf{SAP200} & \textbf{JailbreakChatGPT} & \textbf{T-Jailbreak} & & \textbf{FRR} & \textbf{Score}                      \\ \midrule
\multirow{3}{*}{Llama2-7B-Chat} & n=3             & 1.49          & 0.19          & \textbf{0.00} &                                  5.98         & 1.92 & 7.31 & 3.33 \\
                                & n=5         & 1.30         & 0.25 & \textbf{0.00} & \textbf{5.10}          & 1.66 & 5.79 & 3.25 \\
                                & n=7                    & \textbf{0.74} & \textbf{0.13} & \textbf{0.00} & 5.19          & \textbf{1.52} & \textbf{5.41} & \textbf{3.34} \\
                                \midrule
\multirow{3}{*}{Vicuna-7B}      & n=3             & 14.90          & 2.00          & \textbf{3.95} &                                  \textbf{6.24}         & \textbf{6.77} & \textbf{4.01} & \textbf{3.50} \\
                                & n=5         & \textbf{14.15}          & 2.25 & 5.26 & 7.48          & 7.29 & 5.72 & 3.44 \\
                                & n=7                    & 15.83 & \textbf{1.75} & 6.58 & 7.56          & 7.93 & 4.01 & 3.44 \\
                                \midrule
\multirow{3}{*}{Vicuna-13B}     & n=3             & \textbf{14.15}          & \textbf{4.63}          & \textbf{1.32} &                                  16.53         & \textbf{9.16} & 1.83 & 3.70 \\
                                & n=5         & 14.90          & 5.50 & \textbf{1.32} & \textbf{16.45}          & 9.54 & \textbf{1.64} & \textbf{3.76} \\
                                & n=7                    & 18.25 & 5.75 & \textbf{1.32} & 19.00          & 11.08 & 1.83 & 3.72 \\
                                \bottomrule
 \end{tabular}
 }
 \caption{The ASR ($\%$) results and the helpfulness for different maximum number of guidelines. If the maximum is 7, it means that the number of guidelines provided by GuidelineLLM must be less than or equal to 7. The best results are highlighted in bold.}
 \label{tab:max_number_harmfulness_and_helpfulness}
\end{table*}

To examine the impact of the number of guidelines provided by GuidelineLLM on the harmfulness and helpfulness of the responding LLMs' outputs, we set three different maximum numbers of guidelines: 3, 5, and 7.

Table~\ref{tab:max_number_harmfulness_and_helpfulness} presents the harmfulness results for varying maximum numbers of provided guidelines. For Llama2-7B-Chat, which features more robust alignment mechanisms, we observe that an increase in the number of guidelines provided results in a lower average ASR, reaching as low as 1.52\%. Conversely, for Vicuna-7B and Vicuna-13B, which possess weaker alignment mechanisms, fewer guidelines lead to a lower average ASR, at 6.77\% and 9.16\%, respectively. This indicates that \textbf{the effectiveness of GuidelineLLM is contingent upon the alignment mechanisms of the LLM}. For LLMs with robust alignment mechanisms, supplying detailed guidelines aids in quickly identifying harmful content and subsequently refusing to respond.

Table~\ref{tab:max_number_harmfulness_and_helpfulness} displays the helpfulness results for different maximum numbers of provided guidelines. The data reveal that variations in the number of guidelines have negligible impact on the helpfulness of all three LLMs.

\section{Conclusion}

This study introduces a defensive paradigm called GuidelineLLM, aimed at enhancing the safety of various LLMs. Before responding to a query, GuidelineLLM first identifies potential risks, summarizes them into guideline suggestions, and feeds these to the responding LLMs. This approach reduces the harmfulness of the resulting content without requiring additional safety training for the LLMs. Additionally, we present a fine-tuning framework for GuidelineLLM, which includes the templated construction of jailbreak attacks, referred to as T-Jailbreak. Experiments demonstrate that GuidelineLLM can significantly reduce the harmfulness of output (achieving an average reduction of 34.17\% in the ASR) while maintaining the helpfulness of LLMs in responding to benign queries.

\section{Acknowledgments}
We want to thank all the anonymous reviewers for their valuable comments. This work was supported by National Natural Science Foundation of China (U23B2055, 62376075, 62276077, and 62406188).

\bibliography{aaai25}

\end{document}